\documentclass[conference]{IEEEtran}
\IEEEoverridecommandlockouts
\usepackage{cite}
\usepackage{amsmath,amssymb,amsfonts}
\usepackage{algorithmic}
\usepackage{graphicx}
\usepackage{textcomp}
\usepackage{xcolor}
\def\BibTeX{{\rm B\kern-.05em{\sc i\kern-.025em b}\kern-.08em
    T\kern-.1667em\lower.7ex\hbox{E}\kern-.125emX}}
\begin{document}

\title{Shorten Spatial-spectral RNN with Parallel-GRU for Hyperspectral Image Classification\\
{\footnotesize}
}

\author{\IEEEauthorblockN{Haowen Luo}
\IEEEauthorblockA{\textit{School of Geography and Planning} \\
\textit{Sun Yat-sen University}\\
Guangzhou, China \\
luohw3@mail2.sysu.edu.cn}
}

\maketitle

\begin{abstract}
    Convolutional neural networks (CNNs) attained a good performance in hyperspectral
    sensing image (HSI) classification, but CNNs consider spectra as orderless vectors.
    Therefore, considering the spectra as sequences, recurrent neural networks (RNNs)
    have been applied in HSI classification, for RNNs is skilled at dealing with
    sequential data. However, for a long-sequence task, RNNs is difficult for training
    and not as effective as we expected. Besides, spatial contextual features are not
    considered in RNNs. In this study, we propose a Shorten Spatial-spectral RNN with
    Parallel-GRU (St-SS-pGRU\footnote{The sources are now available at: 
    
    https://github.com/codeRimoe/DL\_for\_RSIs/tree/master/StSSpGRU})  for HSI classification.
    A shorten RNN is more efficient and easier for training than band-by-band RNN.
    By combining converlusion layer, the St-SSpGRU model considers not only spectral
    but also spatial feature, which results in a better performance. An architecture
    named parallel-GRU is also proposed and applied in St-SS-pGRU. With this architecture,
    the model gets a better performance and is more robust.         
\end{abstract}

\begin{IEEEkeywords}
 deep learning,
 gated recurrent unit (GRU),
 long short-term memory (LSTM),
 recurrent neural networks (RNN),
 hyperspectral image classification
\end{IEEEkeywords}

\section{Introduction}
Hyperspectral image (HSI) has attracted considerable attention in the remote sensing
community and been widely used in various areas \cite{bioucas2013hyperspectral}. With
the rich spectral information in HSI, different land cover categories can potentially
be differentiated precisely.

In recent years, deep learning has been widely used in various fields, including HSI
classification \cite{zhu2017deep}. Convolutional neural networks (CNNs) and residual
networks (ResNets) have obtained a successful result for HSI classification
\cite{lee2017going, zhong2017deep}. Recurrent neural networks (RNNs) are also applied
in HSI classification \cite{mou2017deep}.

Because of the ability to extract the spatial contextual information, CNNs and ResNets
can achieve a high accuracy in the classification task. However, CNNs and ResNets
consider spectra as orderless vectors in $d$-dimensional feature space where $d$
represents the number of bands. However, spectra can be seen as orderly and continuing
sequences in the spectral space. In other words, CNNs and ResNets ignore the continuity
of spectra \cite{mou2017deep}.

RNNs have proved effective in solving many challenging problems involving sequential
data, such as Natural Language Processing (NLP) \cite{sundermeyer2015feedforward} and
prediction of time series \cite{gers2002applying}. Considering the spectrum as a
sequential sequence, the application of RNNs is reasonable as it can take full
advantage of the high spectral resolution characteristics of HSI. However, for a
long-sequence task, RNNs is not as effective as we expected. Long distance dependence,
gradient vanish and overfitting are prone to occur \cite{bengio1994learning}. Even if
the long short-term memory network (LSTM) \cite{williams1989learning} is used to solve
the long-distance dependence problem, RNNs is still hard for training and easily
overfitting in a long-sequence task.

In previous work, 3D-CNN is applied in HSI classification and obtained a good behavior
\cite{li2017spectral, zhong2018spectral}. For RNNs, Convolutional-LSTM (CLSTM)
\cite{xingjian2015convolutional} also achieved a good performance in HSI classification
\cite{liu2017bidirectional}. 3D-CNNs and CLSTM consider both spatial contextual
information and spectral continuity, which result in a high accuracy. Nevertheless,
it takes a long time to train these two models.

In \cite{mou2017deep}, LSTM and its variant, GRU \cite{yao2015depth}, are applied in
HIS classification, and it is proved that GRU has a better performance in HIS
classification. To solve the problem that RNNs are easily over-fitting and difficult
for training, \cite{xu2017band} proposed band-group LSTM, which can effectively make
training easier by reducing the number of timestep in LSTM.

In this study, a Shorten Spatial-spectral RNN with Parallel-GRU (St-SS-pGRU) is
proposed. This study contributes to the literature in 2 major respects:

\begin{description}
    \item[1)]
    A shorten RNN with GRU is applied in HIS classification. The model is more
    efficient and easier for training than band-by-band RNN. By combining converlusion
    layer, an advanced model Shorten Spatial-spectral RNN with GRU is proposed.
    The model considers not only spectral but also spatial feature, which leads to a
    better performance.
    \item[2)] 
    An architecture named parallel-GRU is proposed and the model with this architecture
    has a better performance and is more robust.
\end{description}

The remainder of this paper is organized as follows. In the methodology section,
firstly the structure of traditional RNN, LSTM and GRU are introduced and then the
architecture of the proposed models are described. In the experimental section, the
network setup, the experimental results, and the comparison of different models are
provided. Finally, the conclusion section concludes the paper.

\section{Methodology}

\subsection{Recurrent neural networks (RNN)}

Different from Artificial neural network (ANN), RNN \cite{williams1989learning},
a neural network with recurrent unit, has a better performance in solving many
challenging problems involving sequential data analysis. The state of each time
step of the recurrent unit is not only related to the input of the current step,
but also related to the state of the previous step. Thus, the state of the preceding
step can effectively influence the next step.

Given a sequence sample
{$\mathbf{x}={\begin{bmatrix}\mathbf{x}^{(1)},\mathbf{x}^{(2)},...,\mathbf{x}^{(m)}\end{bmatrix}}^{\top}$},
in which {$\mathbf{x}^{(t)}$} is the data at {$t$}th timestep. For the {$t$}th recurrent
unit, its hidden state can be described as:

\begin{equation} 
   \mathbf{h}^{(t)}=\left\{
    \begin{matrix}
       \mathbf{h}^{(0)} & t=0\\ 
       h(\mathbf{h}^{(t-1)},\mathbf{x}^{(t)}) & t>0
    \end{matrix}\right.,
       \label{con:rnnu_h}
\end{equation}
where {$\mathbf{h}^{(0)}$} is the initial state of the recurrent unit, {$h$} is a
nonlinear function. Normally, {$\mathbf{h}^{(0)}$} is set as a zero vector.

Optionally, in $t$th timestep, the recurrent unit may have an output
$\mathbf{y}^{(t)}$. For some task, the RNN model will finally have an output vector
{$\mathbf{y}={\begin{bmatrix}\mathbf{y}^{(1)},\mathbf{y}^{(2)},...,\mathbf{y}^{(m)}
\end{bmatrix}}^{\top}$}, while for classification tasks, only one output is needed.
Generally, The last output is adopted:
\begin{equation} 
    \mathbf{y}^{(t)}=y(\mathbf{h}^{(t)})
    \label{con:rnnu_y}
\end{equation}

\begin{figure}[htbp]
    \centerline{\includegraphics[width=6.12cm]{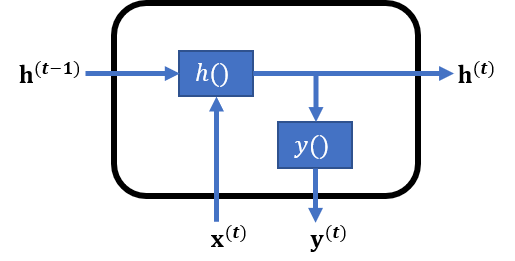}}
    \caption{Graphic model of traditional recurrent unit.}
    \label{fig:rnnu}
\end{figure}

The recurrent unit in a traditional RNN is shown in Fig. \ref{fig:rnnu}. In the
traditional RNN model, the update rule of the recurrent hidden state and output in
Eq. \eqref{con:rnnu_h} and \eqref{con:rnnu_y} is usually implemented as follows:
\begin{equation} 
    h(\mathbf{h}^{(t-1)},\mathbf{x}^{(t)})= \varphi (\mathbf{W}_h\mathbf{x}^{(t)}+\mathbf{U}_h\mathbf{h}^{(t-1)}+\mathbf{b}_h) ,
    \label{con:rnn_tra_h}
\end{equation}
\begin{equation} 
    y(\mathbf{h}^{(t)})= \mathbf{W}_y\mathbf{h}^{(t)}+\mathbf{b}_y ,
    \label{con:rnn_tra_y}
\end{equation}
where $\mathbf{W}_h$, $\mathbf{U}_h$ and $\mathbf{W}_y$ are the weight matrices.
$\mathbf{b}_h$ and $\mathbf{b}_y$ are the bias vectors, and $\varphi$ is an activation
function, such as the sigmoid function or the hyperbolic tangent function.

\subsection{Long short-term memory (LSTM)}

The traditional RNN has the problem of long-distance dependence
\cite{bengio1994learning}. The RNN has the capability to connect different timesteps
related information. However, when the sequence is too long, the RNN becomes unable to
connect related information as the distance increases, because the information losses
when propagating through multi-time-step recurrent units.

\begin{figure}[htbp]
    \centerline{\includegraphics[width=6.1cm]{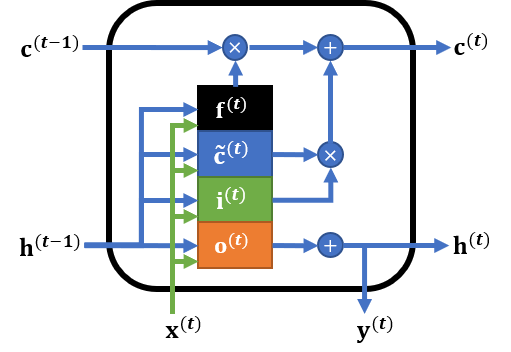}}
    \caption{Graphic model of LSTM.}
    \label{fig:lstm}
\end{figure}

By using long short-term memory (LSTM) \cite{hochreiter1997long}, the problems have
been solved. As Fig. \ref{fig:lstm} shows, LSTM contains a forget gate, an input gate
and an output gate. 'Gate' structure is actually a logistic regression model so that
part of the information is filtered selectively, while the rest is reserved and passes
through the gate. LSTM can simulate the process of forgetting and memory and calculate
the probability of forgetting and memory, so information flow could be preserved in
long-distance propagation. The structure of LSTM can be described as:
\begin{equation}
    \mathbf{f}^{(t)}= \sigma (\mathbf{W}_f\mathbf{x}^{(t)}+\mathbf{U}_f\mathbf{h}^{(t-1)}+\mathbf{b}_f) ,
    \label{con:fg}
\end{equation}
\begin{equation}
    \mathbf{i}^{(t)}= \sigma (\mathbf{W}_i\mathbf{x}^{(t)}+\mathbf{U}_i\mathbf{h}^{(t-1)}+\mathbf{b}_i) ,
    \label{con:ig}
\end{equation}
\begin{equation}
    \mathbf{o}^{(t)}= \sigma (\mathbf{W}_o\mathbf{x}^{(t)}+\mathbf{U}_o\mathbf{h}^{(t-1)}+\mathbf{b}_o) ,
    \label{con:og}
\end{equation}
\begin{equation}
    \tilde{\mathbf{c}}^{(t)}= tanh (\mathbf{W}_c\mathbf{x}^{(t)}+\mathbf{U}_c\mathbf{h}^{(t-1)}+\mathbf{b}_c) ,
\end{equation}
\begin{equation}
    \mathbf{c}^{(t)} = \mathbf{i}^{(t)} * \tilde{\mathbf{c}}^{(t)} + \mathbf{f}^{(t)} * \mathbf{c}^{(t-1)} ,
\end{equation}
\begin{equation}
    \mathbf{h}^{(t)}= \mathbf{o}^{(t)} * tanh(\mathbf{c}^{(t)}),
\end{equation}
where Eq. \eqref{con:fg}, \eqref{con:ig} and \eqref{con:og} represent forget gate,
input gate and output gate.
$\mathbf{W}_f$, $\mathbf{W}_i$, $\mathbf{W}_o$, $\mathbf{W}_c$, $\mathbf{U}_f$,
$\mathbf{U}_i$, $\mathbf{U}_o$ and $\mathbf{U}_c$ are the weight matrices.
$\mathbf{b}_f$, $\mathbf{b}_i$, $\mathbf{b}_o$ and $\mathbf{b}_c$ are the bias vectors.
$\sigma$ refers to sigmoid function and tanh refers to the hyperbolic tangent function:
\begin{equation}
    \sigma(x)=\frac{1}{1+e^(-x)} ,
\end{equation}
\begin{equation}
    tanh(x)=\frac{e^x-e^(-x)}{e^x+e^(-x)} ,
\end{equation}

\subsection{Gated recurrent unit (GRU)}

Over the years, there have been many variants of LSTM, but there is no evidence to show
that there is not a superior variant. Any variant may have advantages in a particular
problem \cite{greff2017lstm}.

GRU \cite{yao2015depth} is a variant of LSTM. With fewer parameters, it is much easier
for training than LSTM, and usually achieves the same performance as LSTM in some tasks
\cite{jozefowicz2015empirical}. It is considered that using GRU in a HSI classification
task is more appropriate than using LSTM \cite{mou2017deep}.

\begin{figure}[htbp]
    \centerline{\includegraphics[width=8cm]{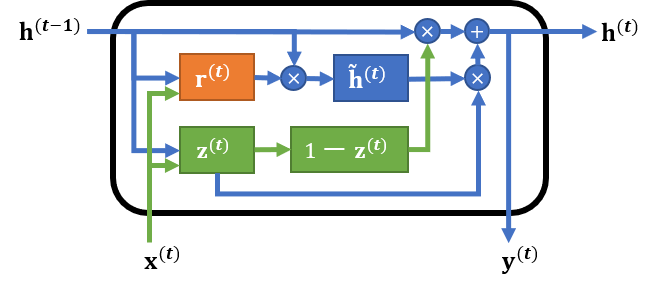}}
    \caption{Graphic model of GRU.}
    \label{fig:gru}
\end{figure}

The main difference between LSTM and GRU is that an update gate and a reset gate are
adopted in GRU, instead of using a forget gate, an input gate and an output gate. The
structure of the GRU is shown in Fig. \ref{fig:gru}, which can be defined as follows:
\begin{equation}
    \mathbf{z}^{(t)}= \sigma (\mathbf{W}_z\mathbf{x}^{(t)}+\mathbf{U}_z\mathbf{h}^{(t-1)}+\mathbf{b}_z) ,
    \label{con:ug}
\end{equation}
\begin{equation}
    \mathbf{r}^{(t)}= \sigma (\mathbf{W}_r\mathbf{x}^{(t)}+\mathbf{U}_r\mathbf{h}^{(t-1)}+\mathbf{b}_r) ,
    \label{con:rg}
\end{equation}
\begin{equation}
    \tilde{\mathbf{h}}^{(t)}= tanh (\mathbf{W}_h\mathbf{x}^{(t)}+\mathbf{U}_h(\mathbf{r}^{(t)}*\mathbf{h}^{(t-1)})+\mathbf{b}_h) ,
\end{equation}
\begin{equation}
    \mathbf{h}^{(t)}= (1-\mathbf{z}^{(t)})\mathbf{h}^{(t-1)} + \mathbf{z}^{(t)}\tilde{\mathbf{h}}^{(t)},
\end{equation}
where Eq. \eqref{con:ug} and \eqref{con:rg} represent update gate and reset gate.
$\mathbf{W}_z$, $\mathbf{W}_r$, $\mathbf{W}_h$, $\mathbf{U}_z$, $\mathbf{U}_r$ and
$\mathbf{U}_h$ are the weight matrices. $\mathbf{b}_z$, $\mathbf{b}_r$ and
$\mathbf{b}_h$ are the bias vectors.

\subsection{The proposed model}

\subsubsection{Shorten Spatial-spectral RNN with GRU(St-SS-GRU)}

\begin{figure}[htbp]
    \centerline{\includegraphics[width=8cm]{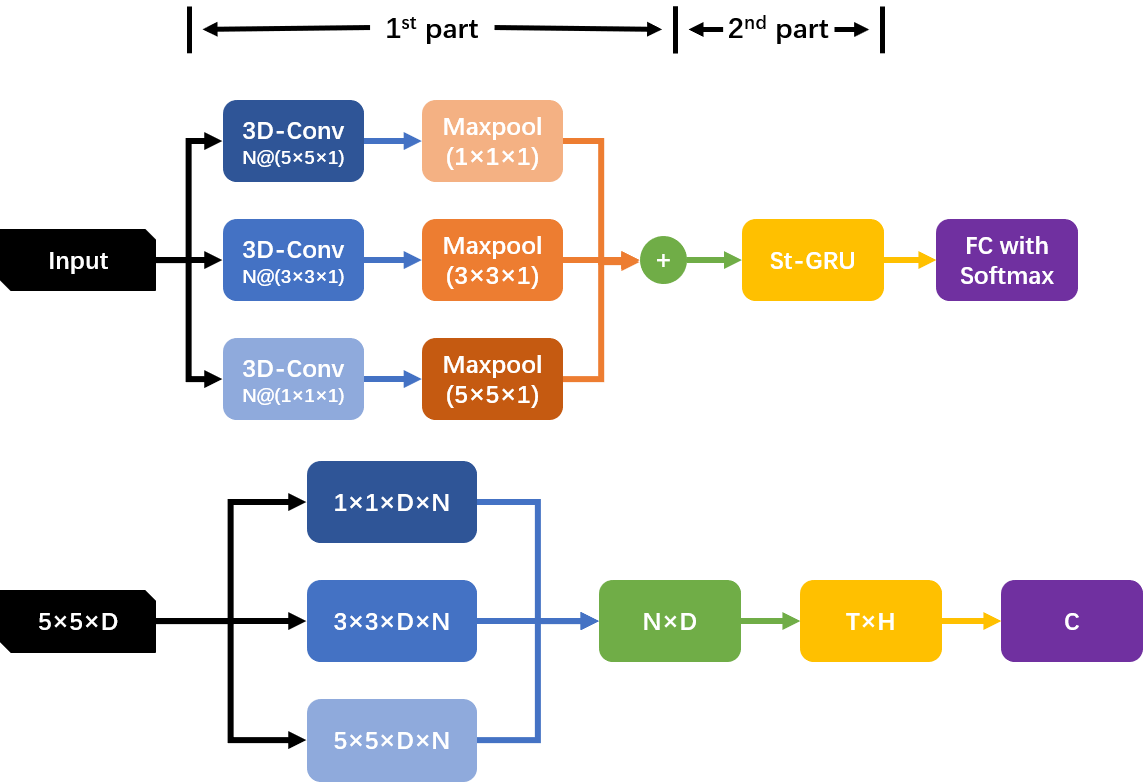}}
    \caption{St-SS-GRU: (1) The first row shows a flowchart of the network. FC refers
    to fully connected layer and Conv refers to Convolutional layers. (2) The second
    row illustrates the shapes of input and output tensors of each layer and their
    connection. (3) N is the number of filters in the 3D-Convolutional layer, D is
    the number of bands in the input image, T is the number of GRU timestep, and H
    is the number of neurons in hidden layer in a GRU. For the Pavia University dataset,
    D=103, and in the experiment the hyperparameters are set as: N=16, T=5, H=128.}
    \label{fig:stssgru}
\end{figure}

A Shorten Spatial-spectral RNN with GRU (St-SS-GRU) model for HSI classification is
shown in Fig. \ref{fig:stssgru}. For each pixel, a square subgraph composed of
5$\times$5 pixels centered on it is used as a training sample.

The first part of St-SS-GRU is actually a 3D-Convolutional layer but both the depth and
stride of the kernels are 1. Three different convolution kernels (1×1, 3×3, 5×5) were
used to convolve different bands. The output of this part is a sequence with the same
length as the original input. The output sequence is a 'spectra' with the spatial
contextual feature. Every timestep of the sequence is a feature vector.

\begin{figure}[htbp]
    \centerline{\includegraphics[width=5cm]{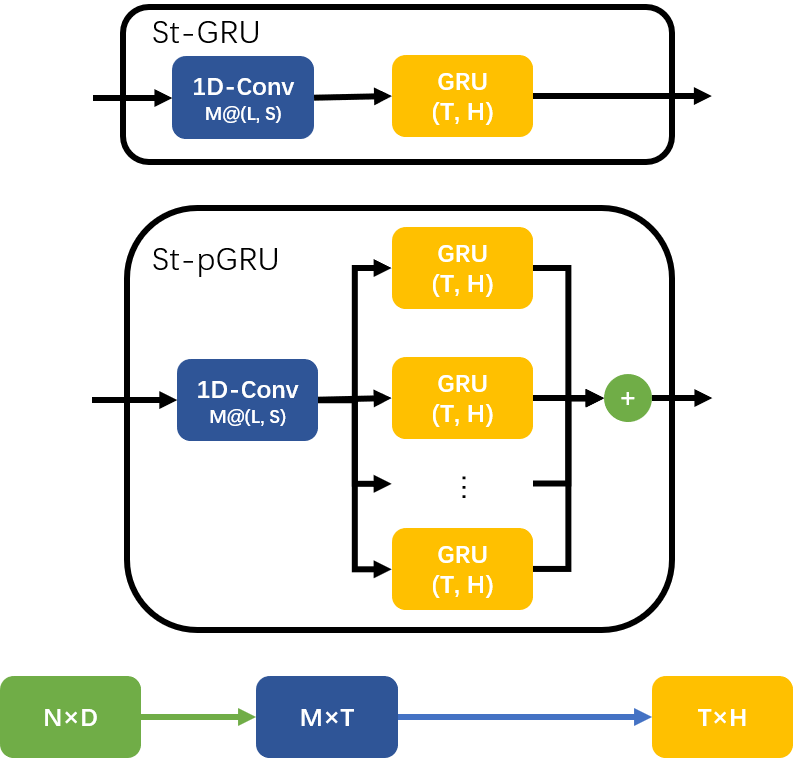}}
    \caption{St-GRU and St-pGRU: (1) The first two rows show the architecture of St-GRU
    and St-pGRU, Conv refers to convolutional layers. (2) The third row illustrates the
    shapes of input and output tensors of each layer and their connection. (3) D is the
    number of bands in the input image, N is the dimension of the vector in each band
    of input, M is the number of filters in the 1D-Convolutional layer, T is the number
    of GRU timestep, and H is the number of neurons in hidden layer in a GRU. L and S,
    which are determined by T, refer to the size and stride of filters in the
    1D-Convolutional layer. For the Pavia University dataset, D=103, and in the
    experiment the hyperparameter is set as: N=M=16, T=5, H=128.}
    \label{fig:stgru}
\end{figure}

The second part is a Shorten RNN with GRU (St-GRU). The structure of St-GRU is shown in
Fig. \ref{fig:stgru}. The 1D converlusion layer before GRU is used to reduce the number
of timesteps so that the network is easier for training.

\subsubsection{Parallel-GRU Architecture}

In order to make the model more robust, a Parallel-GRU (pGRU) architecture is proposed.
The architecture of Shorten Parallel-GRU (St-pGRU) is shown in Fig. \ref{fig:stgru}.
The architecture is actually a combination of several GRU units. The output of the
architecture is the summation of every unit.

The Shorten Spatial-spectral RNN with parallel-GRU (St-SS-pGRU) is similar to St-SS-GRU,
except that St-GRU is replaced by St-SS-pGRU.

\section{Experiment}

\subsection{Data}

In the experiment, two HSI datasets, including the Pavia University and Indian Pines,
are used to evaluate the performance of the proposed model.

The Pavia University dataset was acquired by the Reflective Optics System Imaging
Spectrometer(ROSIS) sensor over Pavia, northern Italy in 2001. The corrected data,
with a spatial resolution of 1.3 m per pixel, contains 103 spectral bands ranging
from 0.43 to 0.86 $\mu m$. The image, with 610$\times$340 pixels, is differentiated into
9 ground truth classes. Table \ref{tab:pusample} provides information about all
classes of the dataset with their corresponding training and test sample.

The Indian Pines dataset was acquired by the TAirborne Visible/Infrared Imaging
Spectrometer (AVIRIS) sensor over the Indian Pines test site in north-western
Indiana in 1992. The corrected data with a moderate spatial resolution of 20m
contains 200 spectral bands ranging from 0.4 to 2.5 $\mu m$. The image consists of
145$\times$145 pixels, which are differentiated into 16 ground truth classes.
Table \ref{tab:ipsample} provides information about all classes of the dataset
with their corresponding training and test sample.

\begin{table}[htbp]
    \caption{Number of Training and Test Samples Used in the Pavia University Dataset}
    \begin{center}
    \begin{tabular}{cc|cc}
    \hline
    \hline
    \textbf{No.} & \textbf{Class Name}& \textbf{Training Samples}& \textbf{Test Samples} \\
    \hline
    1 & Asphalt        &  548 &  6083 \\
    2 & Meadows        &  540 & 18109 \\
    3 & Gravel         &  392 &  1707 \\
    4 & Trees          &  542 &  2522 \\
    5 & Metal sheet    &  256 &  1089 \\
    6 & Bare Soil      &  532 &  4497 \\
    7 & Bitumen        &  375 &   955 \\
    8 & Bricks         &  514 &  3168 \\
    9 & Shadows        &  231 &   716 \\
    \hline
      & \textbf{TOTAL} & 3921 & 38846 \\
    \hline
    \hline
    \end{tabular}
    \end{center}
    \label{tab:pusample}
    \end{table}

    \begin{table}[htbp]
    \caption{Number of Training and Test Samples Used in the Indian Pines Dataset}
    \begin{center}
    \begin{tabular}{cc|cc}
    \hline
    \hline
    \textbf{No.} & \textbf{Class Name}& \textbf{Training Samples}& \textbf{Test Samples} \\
    \hline
    1  & Alfalfa              &   30 &   16 \\
    2  & Corn-notill          &  150 & 1278 \\
    3  & Corn-min             &  150 &  680 \\
    4  & Corn                 &  100 &  137 \\
    5  & Grass-pasture        &  150 &  333 \\
    6  & Grass-trees          &  150 &  580 \\
    7  & Grass-pasture-mowed  &   20 &    8 \\
    8  & Hay-windrowed        &  150 &  328 \\
    9  & Oats                 &   15 &    5 \\
    10 & Soybean-notill       &  150 &  822 \\
    11 & Soybean-mintill      &  150 & 2305 \\
    12 & Soybean-clean        &  150 &  443 \\
    13 & Wheat                &  150 &   55 \\
    14 & Woods                &  150 & 1115 \\
    15 & Building-grass-trees &   50 &  336 \\
    16 & Stone-stell-towers   &   50 &   43 \\
    \hline
       & \textbf{TOTAL}       & 1765 & 8484\\
    \hline
    \hline
    \end{tabular}
    \end{center}
    \label{tab:ipsample}
    \end{table}

\subsection{Result}

\begin{figure}[htbp]\scriptsize
    \centering
     \begin{tabular}{ccc}
     \includegraphics[width=0.14\textwidth]{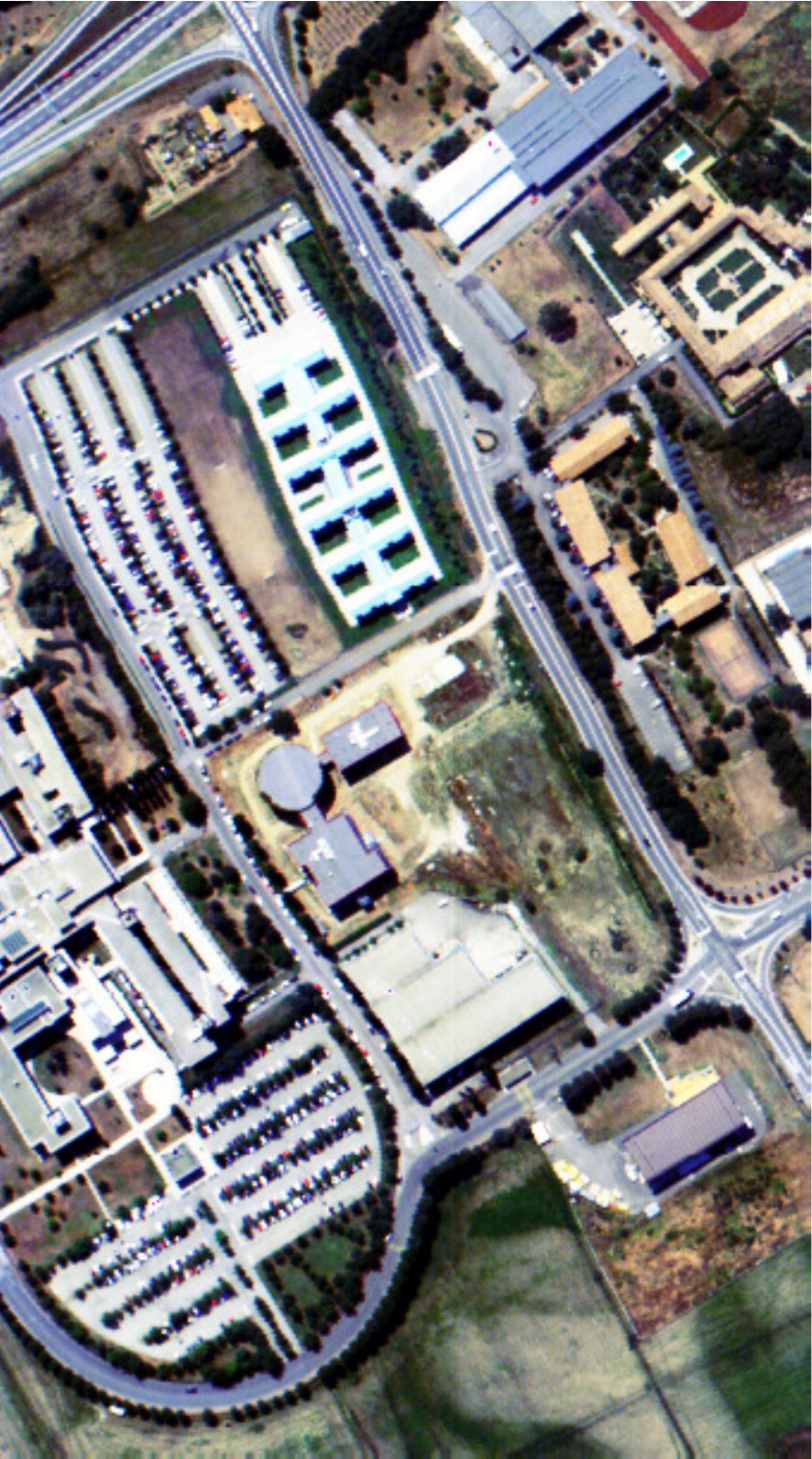}&
     \includegraphics[width=0.14\textwidth]{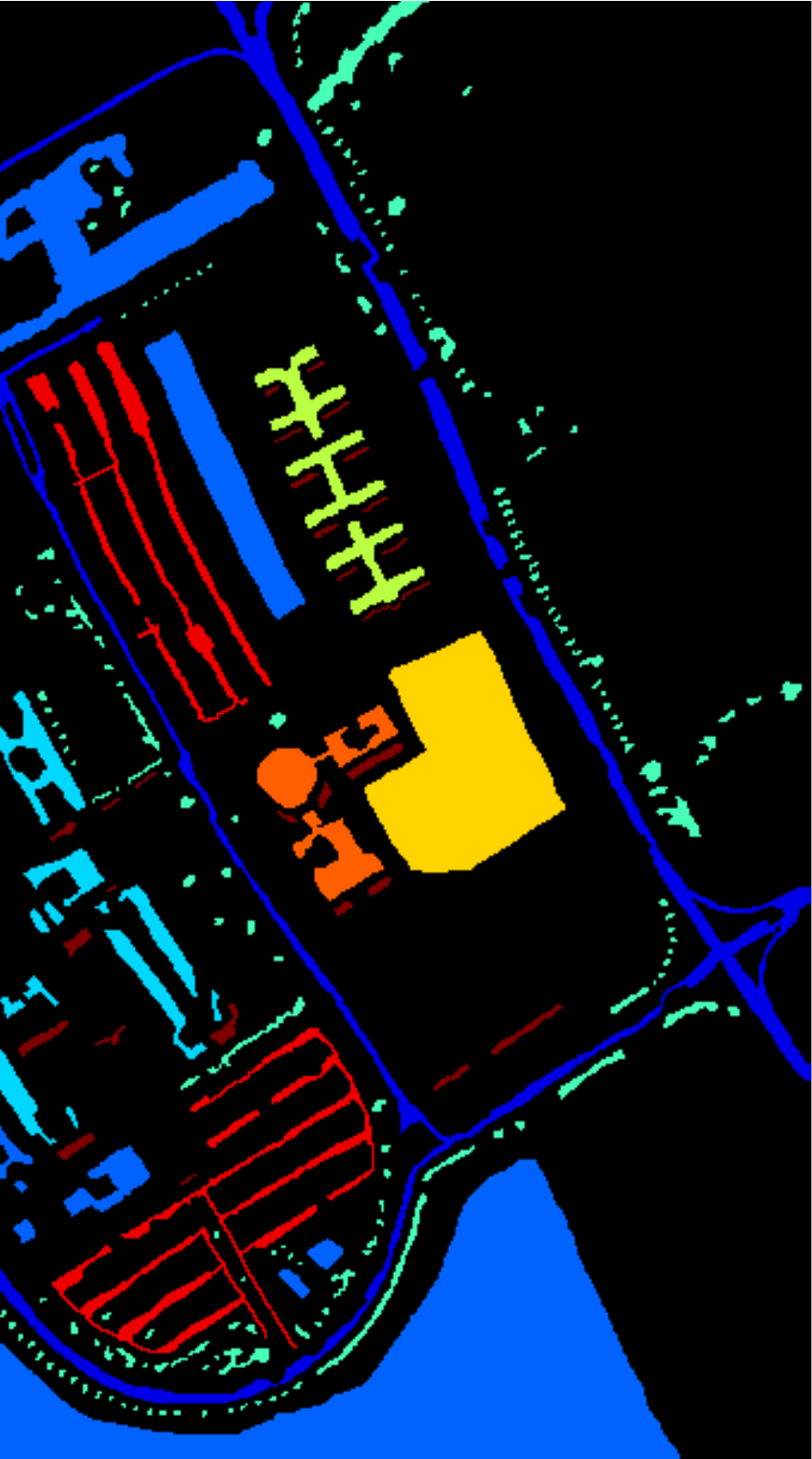}&
     \includegraphics[width=0.14\textwidth]{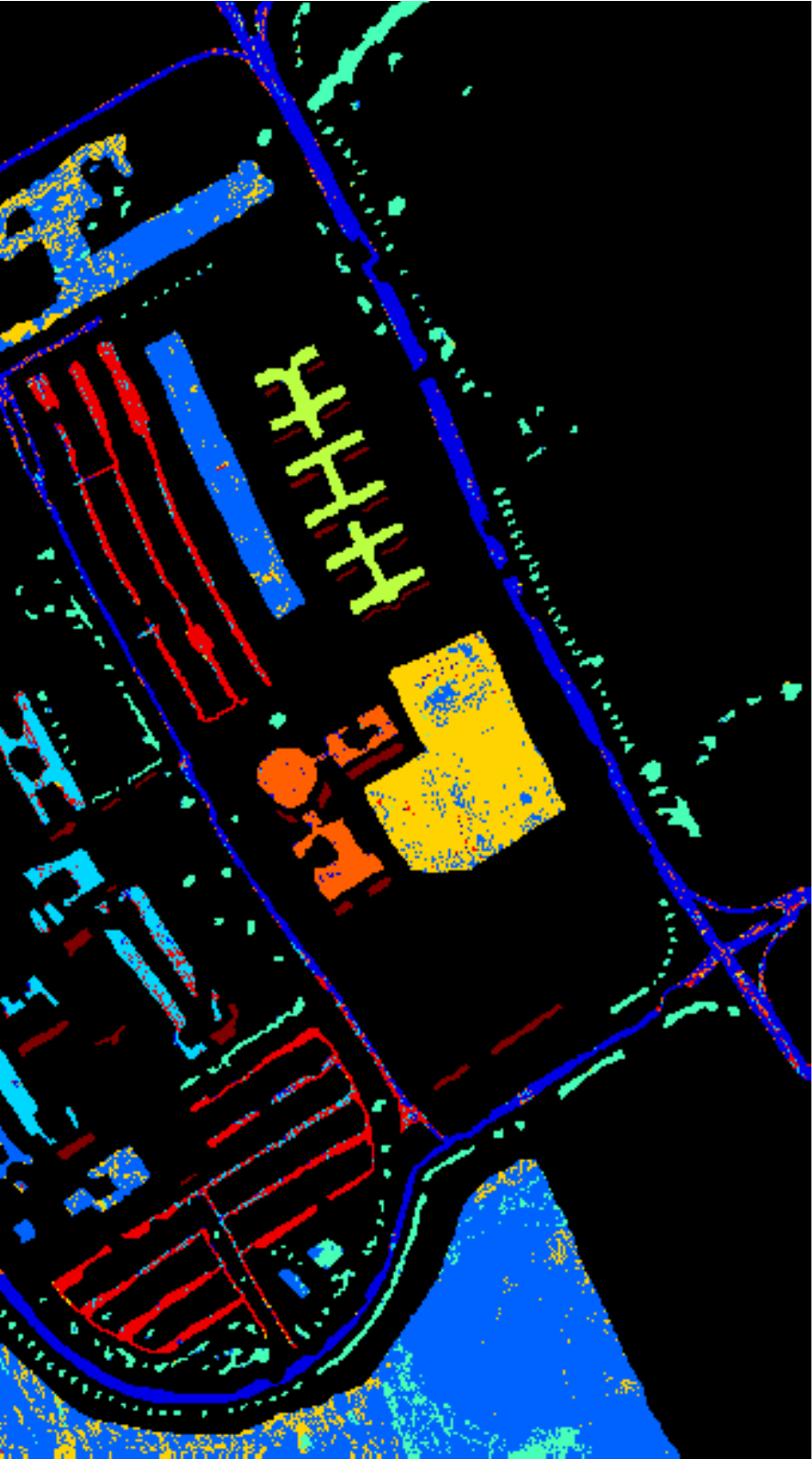}\\
     (a) False color map & (b) Ground truth & (c) GRU(87.79\%)\\
     \includegraphics[width=0.14\textwidth]{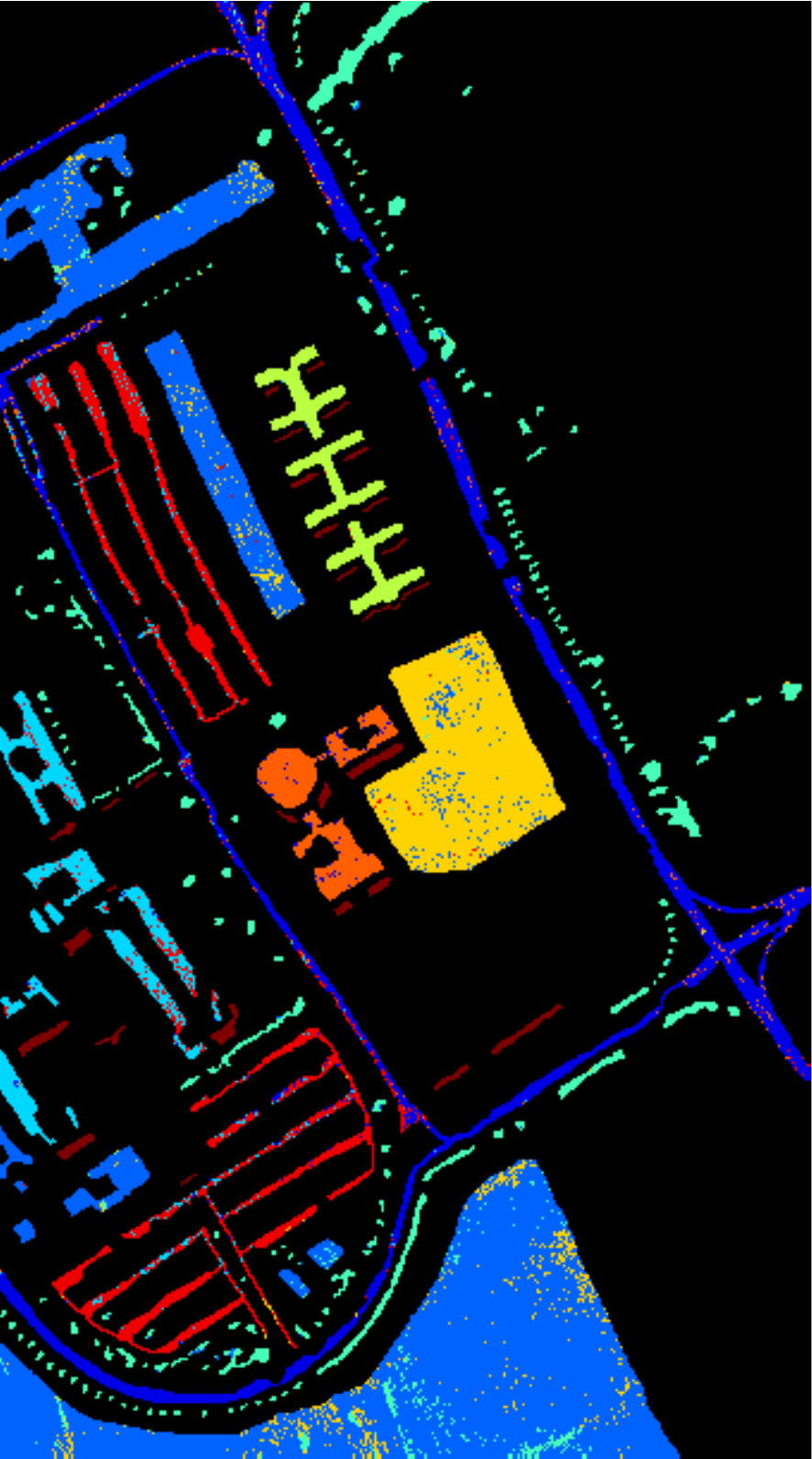}&
     \includegraphics[width=0.14\textwidth]{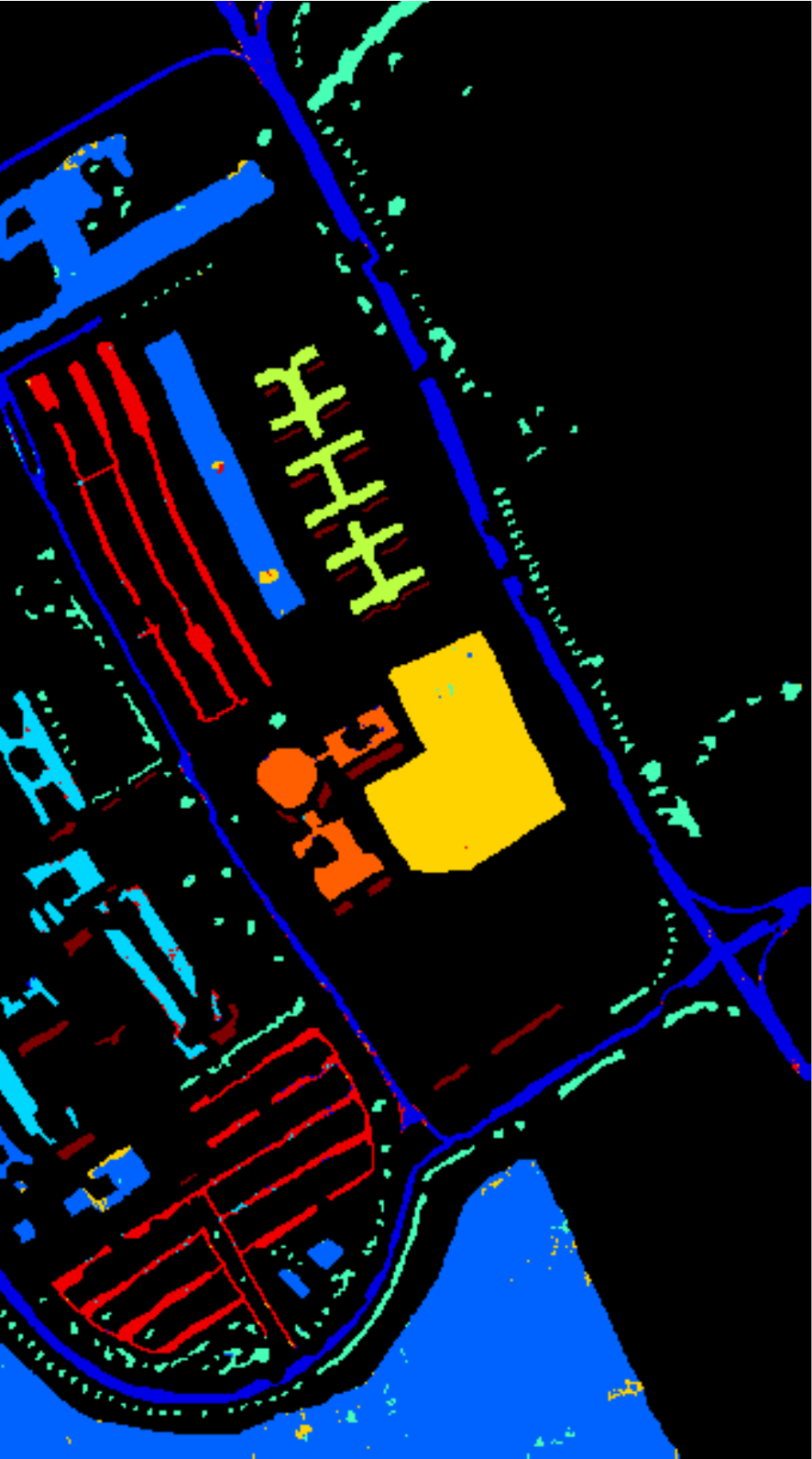}&
     \includegraphics[width=0.14\textwidth]{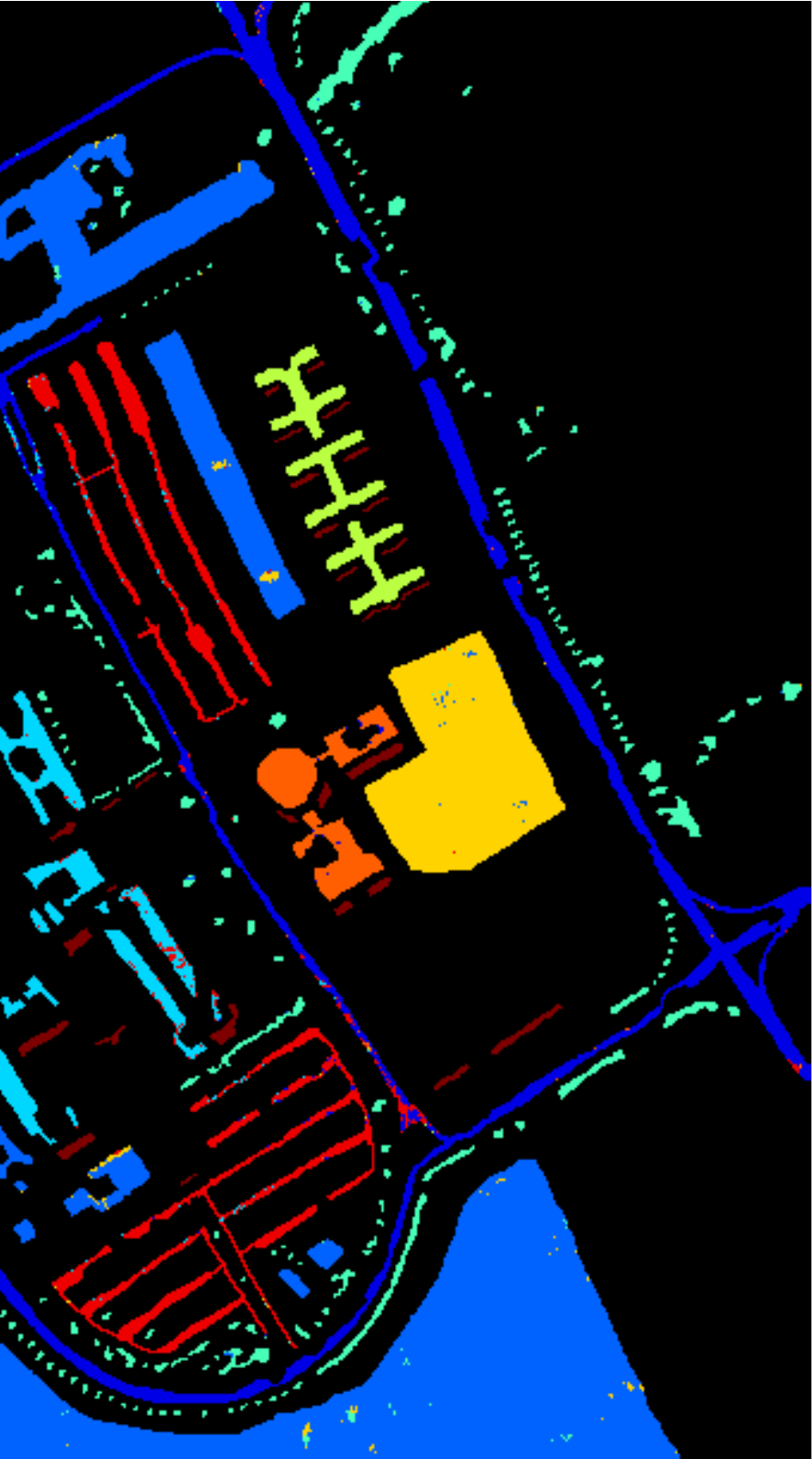}\\
     (e) St-GRU(92.38\%) & (f) St-SS-GRU(98.04\%) & (g) St-SS-pGRU(99.01\%)\\
     \multicolumn{3}{c}{\includegraphics[width=0.42\textwidth]{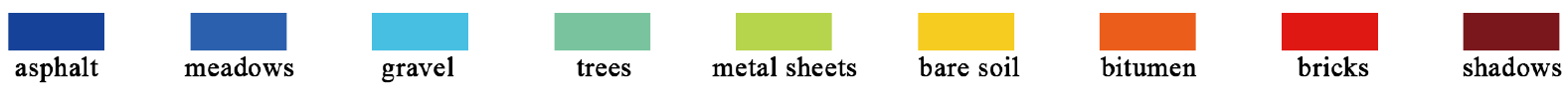}} \\
     \end{tabular}
     \caption{The classification maps of the Pavia University dataset.}
     \label{fig:pumap}
    \end{figure}

    \begin{figure}[htbp]\scriptsize
        \centering
         \begin{tabular}{ccc}
         \includegraphics[width=0.14\textwidth]{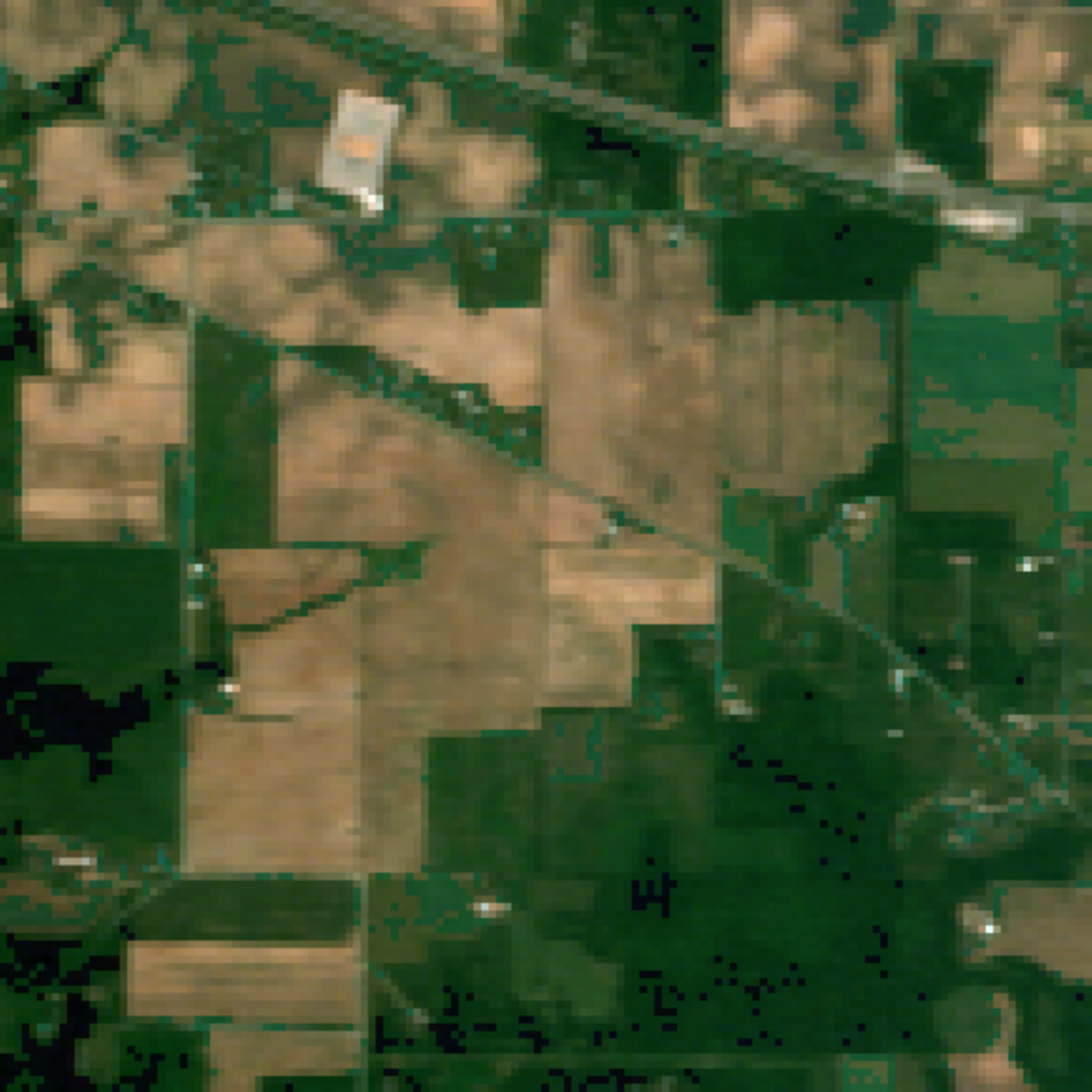}&
         \includegraphics[width=0.14\textwidth]{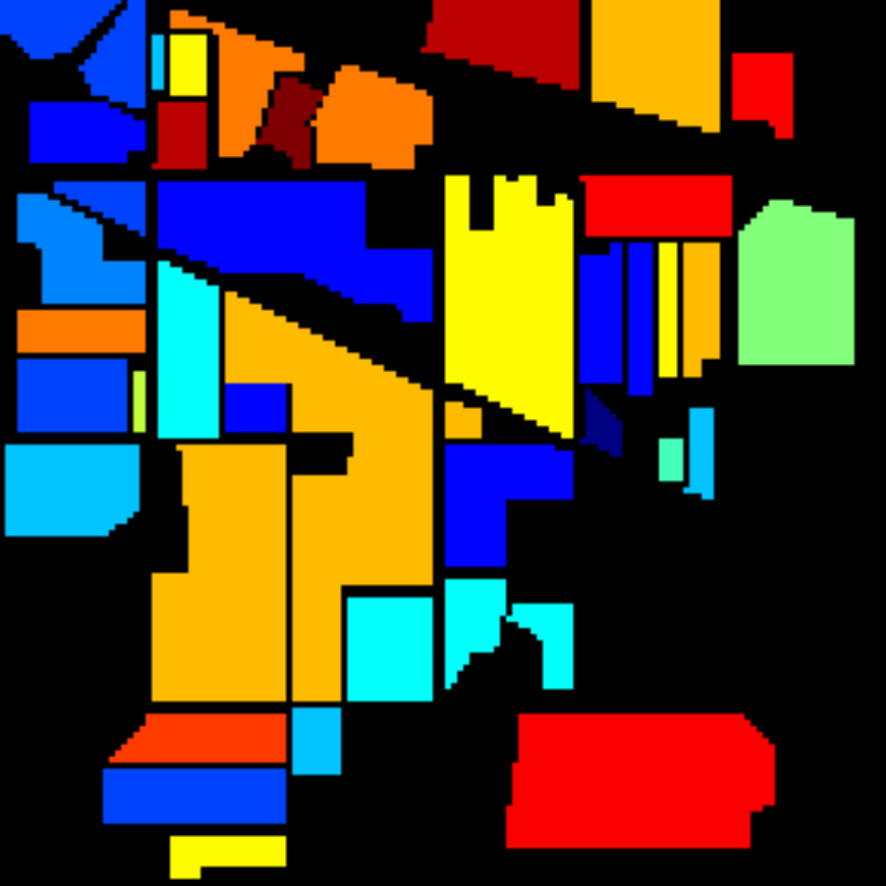}&
         \includegraphics[width=0.14\textwidth]{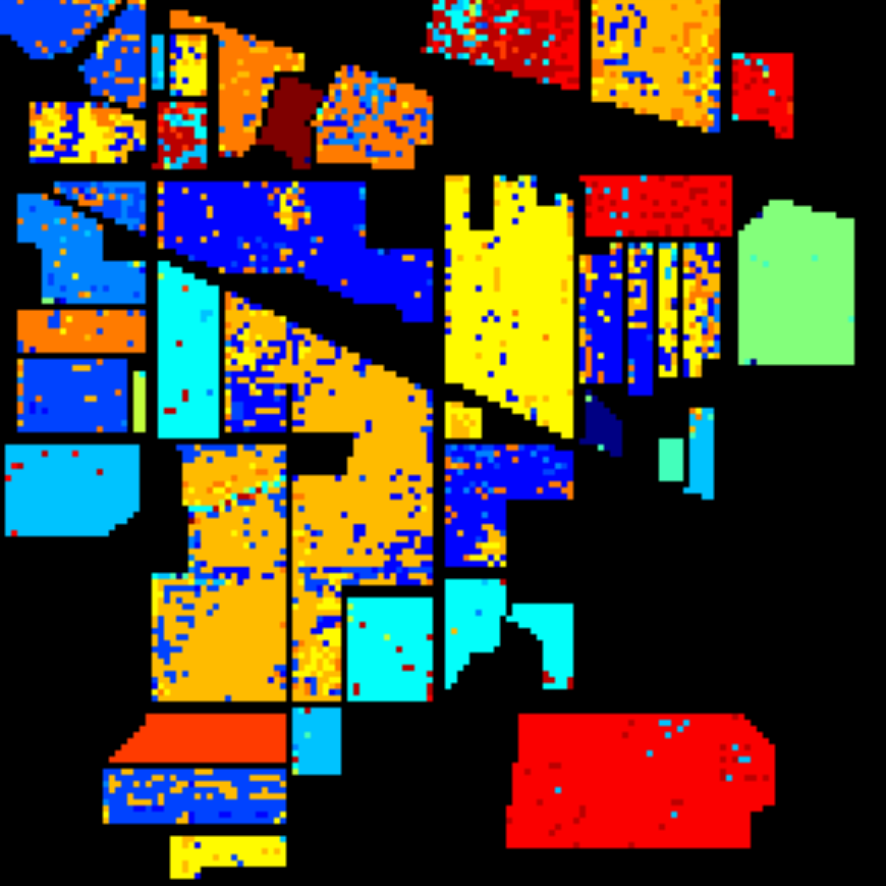}\\
         (a) False color map & (b) Ground truth & (c) GRU(77.07\%)\\
         \includegraphics[width=0.14\textwidth]{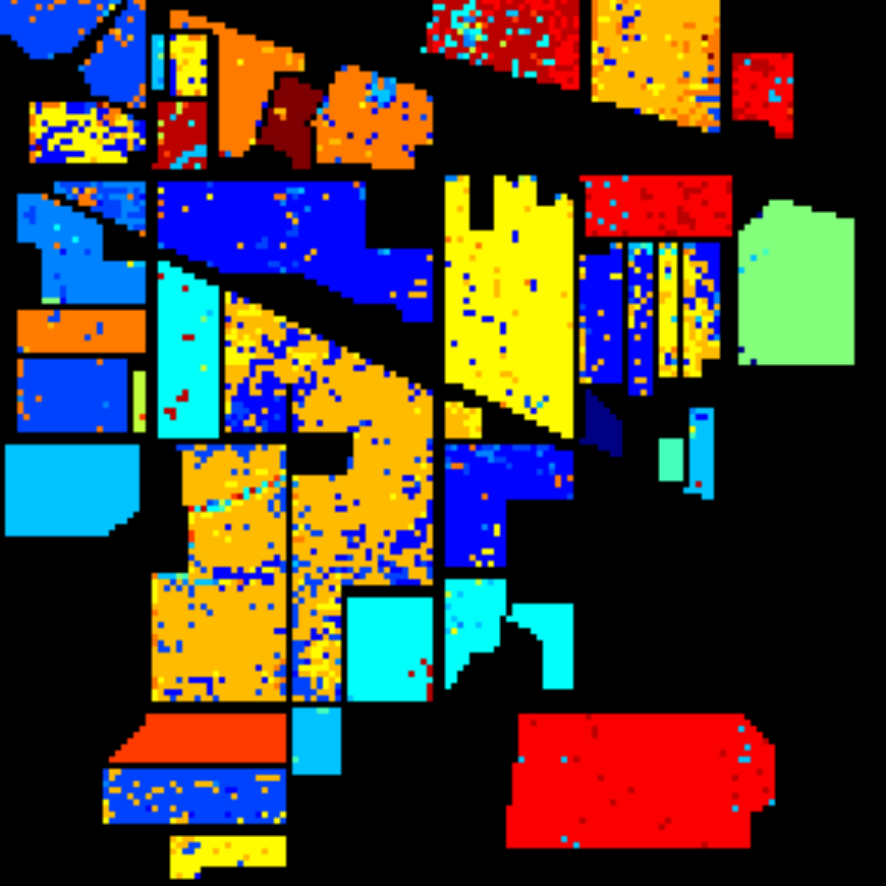}&
         \includegraphics[width=0.14\textwidth]{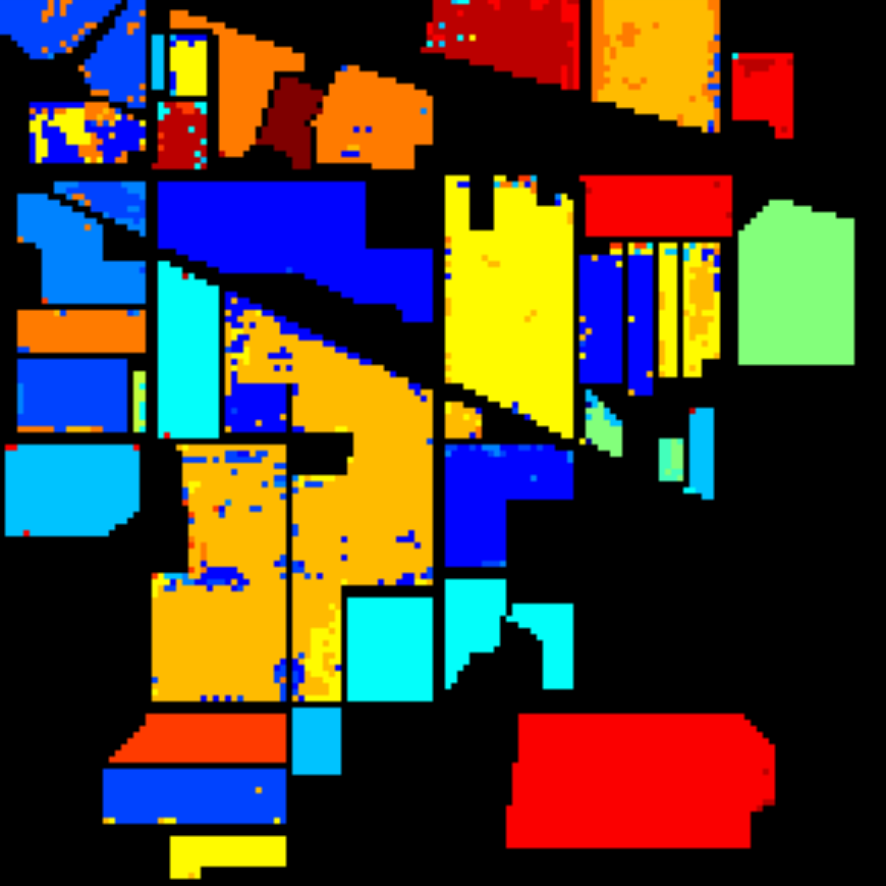}&
         \includegraphics[width=0.14\textwidth]{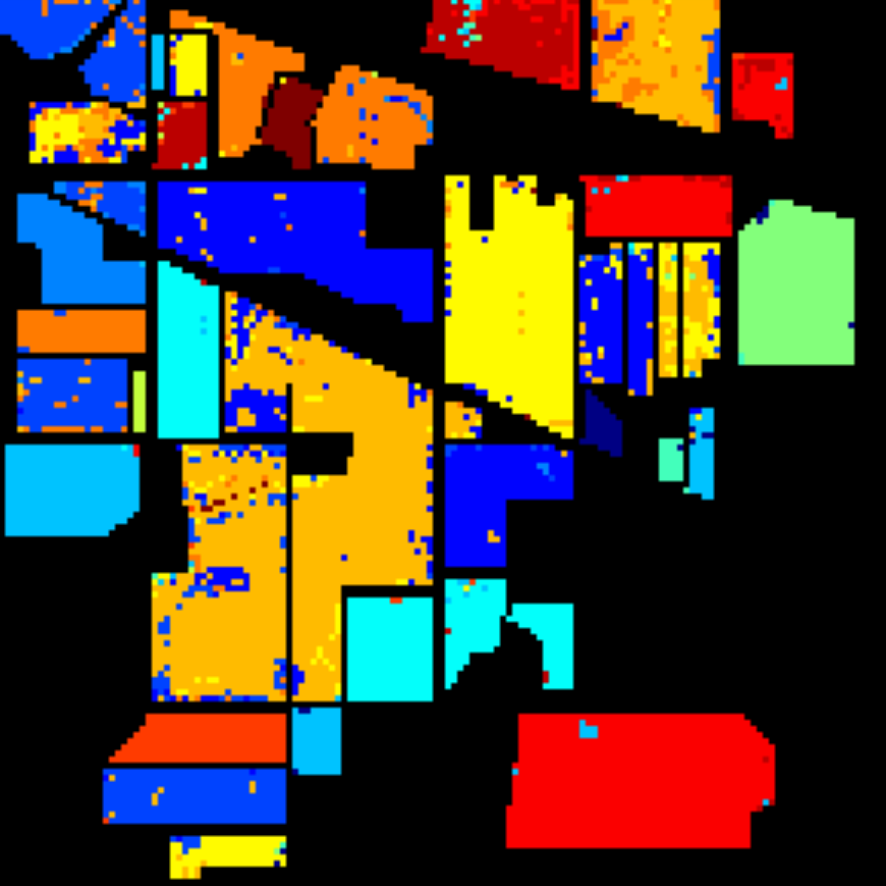}\\
         (e) St-GRU(80.50\%) & (f) St-SS-GRU(86.28\%) & (g) St-SS-pGRU(89.61\%)\\
         \multicolumn{3}{c}{\includegraphics[width=0.42\textwidth]{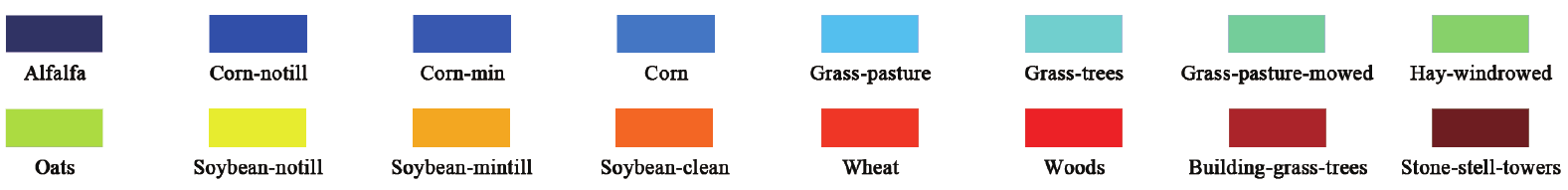}} \\
         \end{tabular}
         \caption{The classification maps of the Indian Pines dataset.}
         \label{fig:ipmap}
        \end{figure}

Table \ref{tab:puaac} and \ref{tab:ipaac} list the results obtained by the experiment,
and Fig. \ref{fig:pumap} and \ref{fig:ipmap} show the classification maps on the Pavia
University dataset and the Indian Pines dataset. Note that the accuracies list in Table
\ref{tab:puaac} and \ref{tab:ipaac} are overall accuracies (OA) along with the standard
deviation, from 10 independent runs on each dataset. The experiment is implemented with
an Intel i7-7700K 4.20GHz processor with 16GB of RAM and an NVIDIA GTX1050Ti graphic
card under Python3.6 with tensorflow1.8.0.

First of all, for all the datasets, GRU outperforms LSTM. In addition, it is observed
that LSTM is difficult to converge in the experiment, while GRU is not. Thus, it is
reasonable to indicate that GRU is a better choice for a HSI classification task.

Furthermore, it is apparent that St-GRU increases the accuracy significantly by 5.33\%
and 3.52\% in the Pavia University dataset and the Indian Pines correspondingly. With
converlusion layers, St-SS-GRU has a better than St-GRU. The accuracy of St-SS-GRU is
4.55\% and 6.63\% higher than that in St-GRU. After parallel-GRU is adopted, the model
gains the best performance in this experiment. The accuracy of St-SS-pGRU is 1.64\%
and 3.19\% higher than St-SS-GRU. What is more, the standard deviation of St-SS-pGRU
is smaller than other models, which indicate that St-SS-pGRU is more robust. 

Comparing the processing time of different methods, st-GRU is significantly faster
in training than band-by-band GRU. St-SS-GRU and St-SS-pGRU are as slow as LSTM and
GRU in training, but they have higher accuracies than LSTM and GRU.

\begin{table}[htbp]
    \caption{Classification Accuracies and Training Time for the Pavia University Dataset}
    \begin{center}
    \begin{tabular}{c|cc}
    \hline
    \hline
    \textbf{Model} & \textbf{Overall accuracy} & \textbf{Training Time (s)} \\
    \hline
    LSTM           & 84.68$\pm$1.40\%               & 434.22                    \\
    GRU            & 86.92$\pm$1.29\%               & 232.15                    \\
    St-GRU         & 92.25$\pm$0.78\%               & \textbf{7.31}*            \\
    St-SS-GRU      & 96.80$\pm$0.37\%               & 104.56                     \\
    St-SS-pGRU     & \textbf{98.44$\pm$0.26\%}*     & 128.91                     \\
    \hline
    \hline
    \multicolumn{3}{l}{* The best performance in each column is shown in bold.}\\
    \end{tabular}
    \end{center}
    \label{tab:puaac}
    \end{table}
    
    \begin{table}[htbp]
    \caption{Classification Accuracies and Training Time for the Indian Pines Dataset}
    \begin{center}
    \begin{tabular}{c|cc}
    \hline
    \hline
    \textbf{Model} & \textbf{Overall accuracy} & \textbf{Training Time (s)}    \\
    \hline
    LSTM           & 71.65$\pm$1.05\%               & 838.85                   \\
    GRU            & 77.01$\pm$1.82\%               & 442.67                   \\
    St-GRU         & 80.53$\pm$0.90\%               & \textbf{7.63}*           \\
    St-SS-GRU      & 87.16$\pm$1.06\%               & 287.54                   \\
    St-SS-pGRU     & \textbf{90.35$\pm$0.86\%}*     & 300.90                   \\
    \hline
    \hline
    \multicolumn{3}{l}{* The best performance in each column is shown in bold.}\\
    \end{tabular}
    \end{center}
    \label{tab:ipaac}
    \end{table}

\section{Conclusion }

In the study, a St-SS-pGRU model is proposed for HSI classification. What is more,
an architecture named parallel-GRU is proposed to promote the performance and
robustness. Then an experiment is conducted to compare the performance of different
models. From the experiment, it is confirmed that GRU performs better than LSTM in
HSI classification task. Moreover, it is apparent that the proposed models are a lot
more accurate, more robust and faster than the traditional GRU network. Specifically,
St-GRU effectively reduced the training time and promoted the accuracy. St-SS-GRU
needs more time for training but gains a better performance than St-GRU. The proposed
architecture parallel-GRU also provided a satisfactory result in the experiment.

\bibliographystyle{unsrt}
\bibliography{refs}
\vspace{12pt}

\end{document}